\documentclass[10pt,twocolumn,letterpaper]{article}

\usepackage{times}
\usepackage{epsfig}
\usepackage{graphicx}
\usepackage{amsmath}
\usepackage{amssymb}
\usepackage{floatrow}
\usepackage{float, csquotes}

\newcommand{\ADDRESSED}[1]{{}}

\newcommand{\hidden}[1]{{}}

\renewcommand{\paragraph}[1]{{\textbf{#1.}}}

\begin{document}


\title{Dynamic Edge Weights in Graph Neural Networks for 3D Object Detection}

\author{Sumesh Thakur and Jiju Peethambaran\\
Graphics and Spatial Computing Lab\\
Saint Mary's University, Halifax, NS, Canada\\
{\tt\small sumesh.thakur@smu.ca, jiju.poovvancheri@smu.ca}
}

\maketitle

\begin{abstract}
 A robust and accurate 3D detection system is an integral part of autonomous vehicles. Traditionally, a majority of 3D object detection algorithms focus on processing 3D point clouds using voxel grids or bird’s eye view (BEV). Recent works, however, demonstrate the utilization of the graph neural network (GNN) as a promising approach to 3D object detection. In this work, we propose an attention based feature aggregation technique in GNN for detecting objects in LiDAR scan. We first employ a distance-aware downsampling scheme that not only enhances the algorithmic performance but also retains maximum geometric features of objects even if they lie far from the sensor. In each layer of the GNN, apart from the linear transformation which maps the per node input features to the corresponding higher level features, a per node masked attention by specifying different weights to different nodes in its first ring neighborhood is also performed. The masked attention implicitly accounts for the underlying neighborhood graph structure of every node and also eliminates the need of costly matrix operations thereby improving the detection accuracy without compromising the performance. The experiments on KITTI dataset show that our method yields comparable results for 3D object detection.

\end{abstract}

\section{Introduction}

Perception of the surrounding environment is crucial for the safe functioning of an autonomous mobile robot or a self driving car. While deployed on the roads, a self driving car utilizes simultaneous localization and mapping (SLAM) algorithm \cite{cadena2016past} to  infer its spatial position relative to other vehicles, pedestrians and cyclists present in its close proximity. Such localization and the understanding of the  objects in the scene is essential for a self driving car to make crucial decisions about braking, turnings and lane changing. A core enabler of the SLAM and scene understanding algorithms for autonomous driving is the object detection. Considering the real-time scenario and the uncertainty with the object movements, the object detection algorithms must be highly accurate and robust, i.e., detection algorithms cannot afford false detections even under varying environmental conditions and significant occlusions.
\begin{figure}[t]
\begin{center}
   \includegraphics[width=0.8\linewidth]{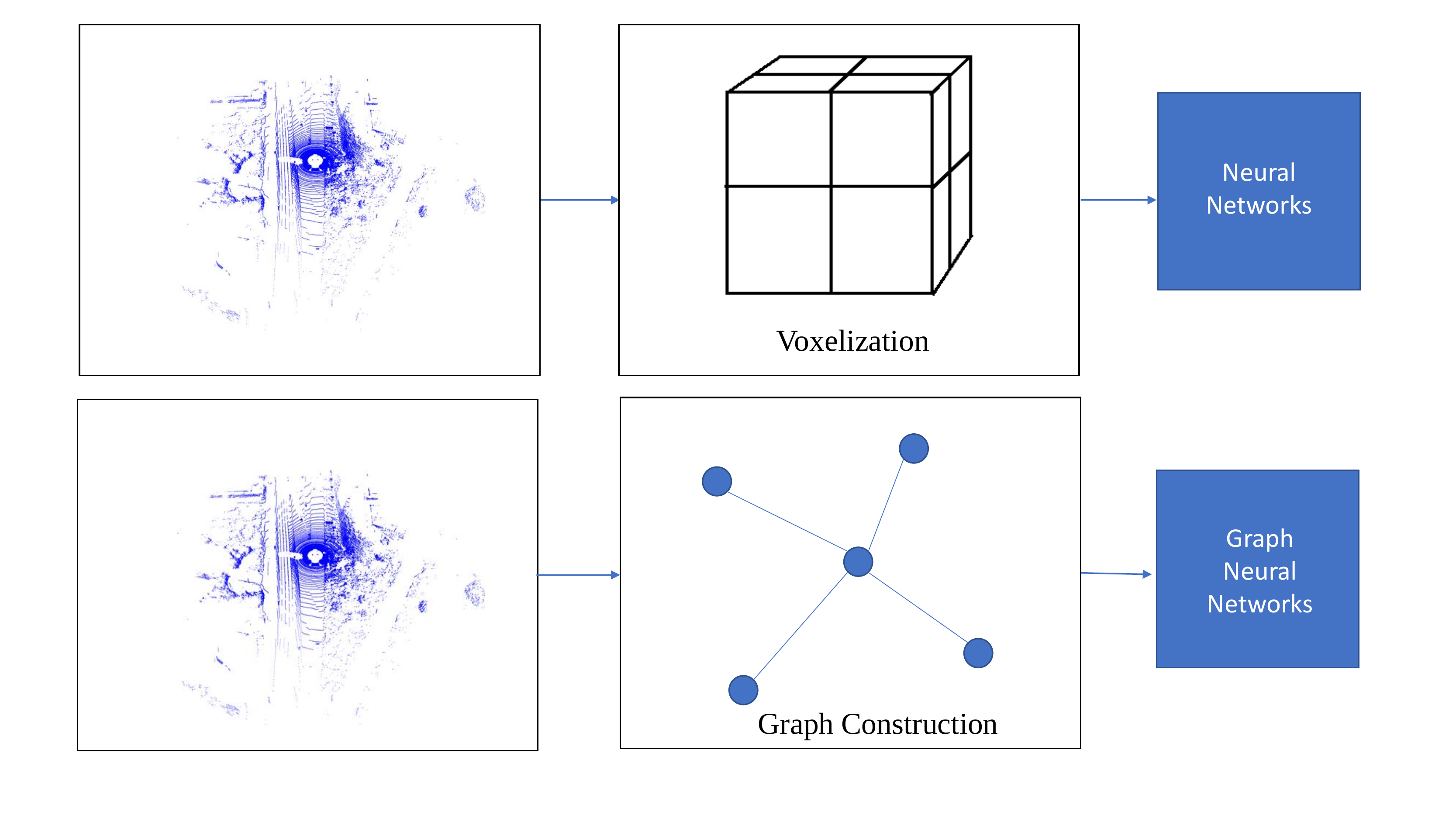}
\end{center}
   \caption{Different representations of point couds.}
\label{fig:long}
\label{fig:onecol}
\end{figure}

Over the last few years, deep learning methods have been shown to outperform state-of-the-art object detection techniques. Modern deep learning based successful object detectors such as Faster R-CNN \cite{ren2015faster}, SSD \cite{liu2016ssd}, R-FCN \cite{dai2016r} and YOLO \cite{redmon2016you} classify objects with 2D bounding boxes. These networks are found to perform well enough to be deployed in consumer products. However, 2D detectors fail to capture all the required information such as depth and orientation from the surrounding environments in 3D. The addition of a third dimension, sparsity of input data and deterioration of its quality as a function of distance all contribute to the failure of 2D object detection methods in 3D. The is quite evident on standard benchmark datasets like KITTI object detection benchmark \cite{geiger2013vision}.

In recent years, several works have been proposed to achieve better results for 3D object detection using point clouds. The most common approaches~\cite{Barrera2020, chen2017multi} perform the object detection task by projecting the input point cloud to different views like bird’s eye view (BEV) or spatially organizing the point clouds into regular equally spaced voxels~\cite{zhou2018voxelnet}. Recently, there has been an increased focus on the generalization of the point clouds to graph representations to perform classification and segmentation tasks.  Some recent methods like~\cite{bi2019graph, landrieu2018large, shen2018mining, wang2019dynamic} have obtained commendable results on object classification and segmentation. A graph neural network (GNN) iteratively updates its vertex features by aggregating features along the edges. The approach is similar to convolution neural networks, however, GNN allows more complex features to be determined along the edges.  So far, only few investigations have been carried out in the direction of designing a GNN based algorithm for object detection, where an optimal prediction of the object shape is required along with its location in environment. PointGNN~\cite{shi2020pointgnn} can be considered as the first GNN based algorithm for 3D object detection in road scenes. Figure \ref{fig:onecol} contrasts the voxel representation with the graph representation of data.

The idea of graph attention networks (GAT) was introduced by Velivckovic et al.~\cite{velivckovic2017graph}. GAT implicitly specify different weights to different vertices in a neighborhood based on their similarity with each other. To assign relative weights, GAT uses concatenation of two neighboring features. Unlike the GAT, we utilize the relative coordinates along with the vertex features to produce relative edge weights. The key idea of our work is as follows. Based on the position and feature attributes of the neighbourhood vertices, we learn to strengthen or weaken the edge weightage accordingly. This approach allows our model to dynamically adapt to the structure of the objects. To enhance the algorithmic performance, we introduce a new distance-aware voxelization method, which downsamples the point cloud scans from LiDAR sensor without loosing the relevant information required for graph generation. Our method is tested on widely used KITTI benchmark dataset. In summary, we make the following key contributions.
\begin{itemize}
  \item \textbf{Down sampling algorithm} :
  We introduce a distance-aware downsampling algorithm, that employs variable sized voxels depending on the distance of points from the sensor origin to subsample the point cloud. The algorithm maximizes the geometric features of objects even if they lie far from the sensor origin.
  \item \textbf{Attention Based Feature Aggregation} :  We design a single stage GNN based algorithm for object detection and localization. Feature aggregation is performed using a masked attention by assigning different weights to different neighboring nodes.
  \item \textbf{Evaluation on KITTI Dataset}: We evaluate the proposed 3D object detection algorithm on the KITTI benchmark and analyze the efficacy of the downsampling and attention mechanisms in an ablation study.
\end{itemize}

\section{Related Work}
Many sub-problems in autonomous driving such as prediction, planning, and motion control require a faithful representation of the 3D space around the autonomous vehicle. Generally this 3D data is captured by LiDAR sensor in the form of point cloud. A point cloud is a set of data points in space where every point in space has $x$, $y$, $z$ values representing its spatial position with respect to the Lidar sensor along with the reflectance value $i$. In 2017, Qi et al. proposed PointNet \cite{pointnet} which showed how to directly manipulate point cloud with neural networks. Since then, most of the research on point cloud processing have been shifted to exploit various 3D data representations using neural networks.

Following the trend of 2D object detectors, the traditional methods for 3D proposal generation utilized hand-crafted features to generate a small set of candidate boxes that retrieve most of the objects in 3D space. 3DOP \cite{chen20153d} is one of the signiﬁcant algorithms in this direction. It uses hand-crafted geometric features from stereo point clouds to score 3D sliding windows in an energy minimization framework. The top k-scoring windows are selected as region proposals, which are then fed to a modiﬁed Fast-RCNN to ﬁnally generate 3D bounding boxes. MONO3D \cite{weng2019monocular} uses the same framework, but instead exploits plane prior and handcrafted features from semantic segmentation outputs to generate 3D proposals from monocular images.

\subsection{Voxel based Methods}
Voxel (also known as volumetric pixel or volume elements) is the smallest unit of volume when dividing 3D space into discrete, uniform regions. Voxel based object detection converts a point cloud data into symmetrical 3D grid and  inputs it into convolution based layers. This approach extends the basic principle of 2D object detectors to 3D object detection. VeloFCN   \cite{li20173d} projects a LIDAR point cloud to the front view, which is fed into a fully convolutional network to generate dense 3D bounding boxes.
Zhou et al. \cite{zhou2018voxelnet} presents a voxel-based end-to-end trainable framework called VoxelNet.  A point cloud is partitioned into equally spaced voxels and per voxel features are encoded into a 4D tensors. Each voxel is further passed through voxel feature extractor (VFE). A region proposal network \cite{ren2015faster} is then utilized to produce the detection results. Although detection results are highly accurate in this case, this method is very slow due to the sparsity of voxels and 3D convolutions.
3D-FCN \cite{DBLP:journals/corr/LiZX16} extends this concept by applying 3D convolutions on 3D voxel grid from LiDAR point clouds to generate better 3D bounding boxes. The limitation of both these methods is that they use very expensive 3D convolution layers, which make it difficult to be deployed in real-time driving scenarios.

\subsection{Point based Methods}
Point based methods process point cloud data directly without representing or projecting it to other views such as front view or bird eye view. These methods model each point independently using several Multi-Layer Perceptron’s (MLPs) and then aggregate a global feature using a symmetric function (e.g., maxpool). These networks can achieve permutation invariance for unordered 3D point clouds. However, the geometric relationships among 3D points are not fully considered. Lang et al. \cite{DBLP:journals/corr/abs-1812-05784} proposed a 3D object detector named Point Pillars. This method leverages PointNet to learn the feature of point clouds organized in vertical columns (called Pillars) and encodes the learned features into a pseudo image. A 2D object detection pipeline is then applied to predict 3D bounding boxes. TA-Net \cite{liu2019tanet} is one of the few deep learning based 3D detection technique that targets pedestrian detection. TA-Net contains a Triple Attention (TA) module, and a Coarse-to-Fine Regression (CFR) module. By considering the channel-wise, point-wise and voxel-wise attention jointly, the TA module enhances the crucial information of the target while suppressing the unstable cloud points. 3D SSD \cite{yang20203dssd} is the latest approach in point based 3D detection. It uses fusion sampling strategy joined with anchor free regression head to regress bounding box co-ordinates. This approach achieves a good balance between accuracy and inference.
\subsection{Graph based Methods}
Recently, Graph neural networks (GCNN) has also been exploited for application in 3D using point clouds. Wang et al. \cite{wang2019dynamic} proposes EdgeConv operation to extract local features in a graph. Bi et al. \cite{bi2019graph} uses GNNs for 3D object classification. PointGNN \cite{shi2020pointgnn} can be regarded as one of the initial methods that uses the graph construction for 3D predictions. It designs a one-stage graph neural network to predict the category and shape of the object with an auto-registration mechanism, PointRGCN \cite{zarzar2019pointrgcn} leverages a graph representation for feature generation. This method uses a 2-stage object detection, where R-GCN is a residual graph CNN that classifies and regress 3D proposals, and  C-GCN a contextual GCN that further refines proposals by sharing contextual information between multiple proposals.

\section{Graph Attention Networks for 3D Object Detection}
\begin{figure*}[h]
\begin{center}
\includegraphics[width=16cm]{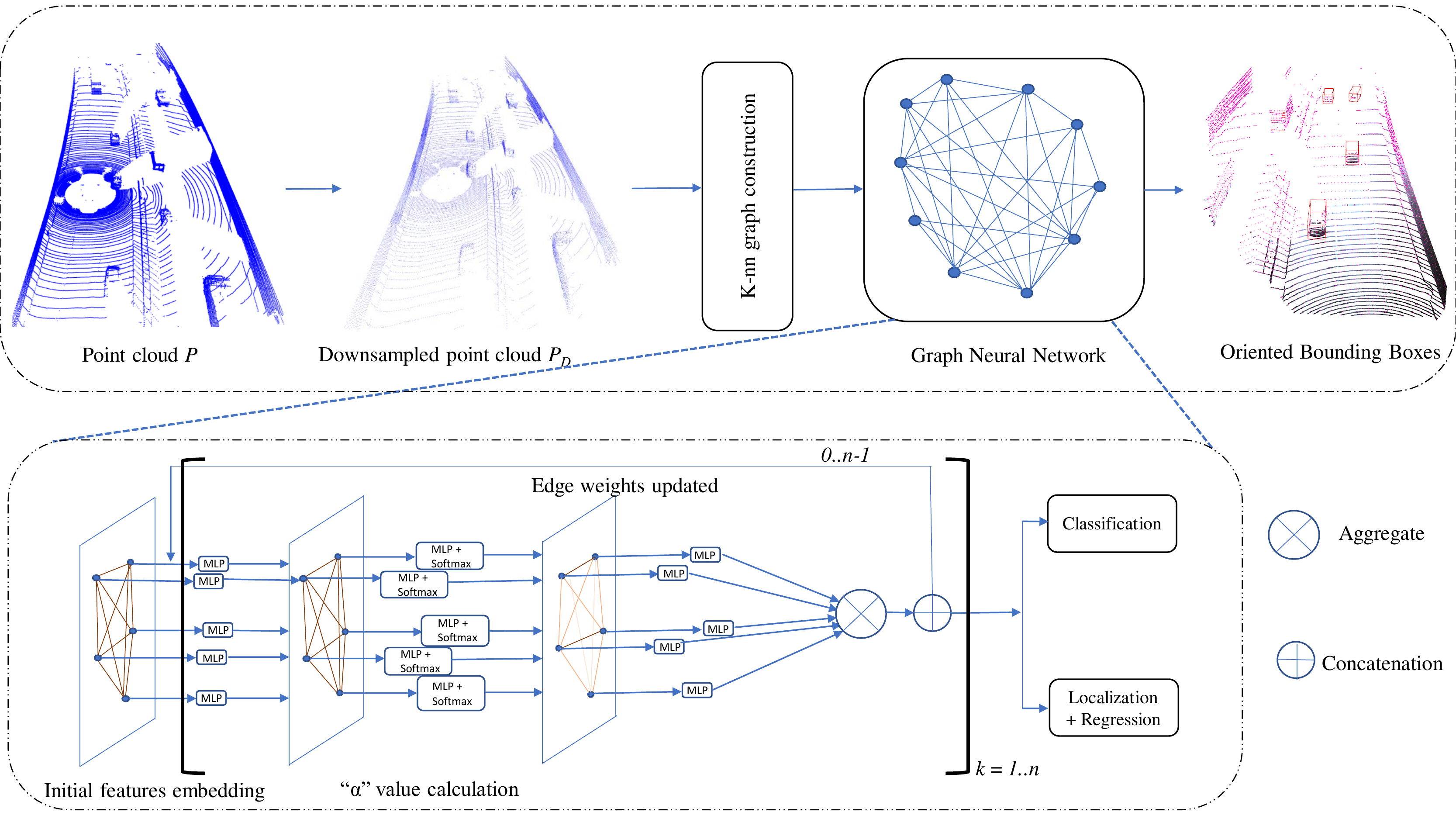}
\end{center}
   \caption{The architecture of the proposed approach. The blocks in the square brackets constitute various steps in one iteration of the proposed GNN. }
\label{fig:architecture}
\end{figure*}

In this section, we discuss the proposed approach to detect 3D objects from LiDAR scans. Our architecture mainly consists of two modules: graph construction and an attention based GNN (refer to Figure \ref{fig:architecture}).

\begin{figure}[h]
\begin{center}
\includegraphics[width=1\linewidth]{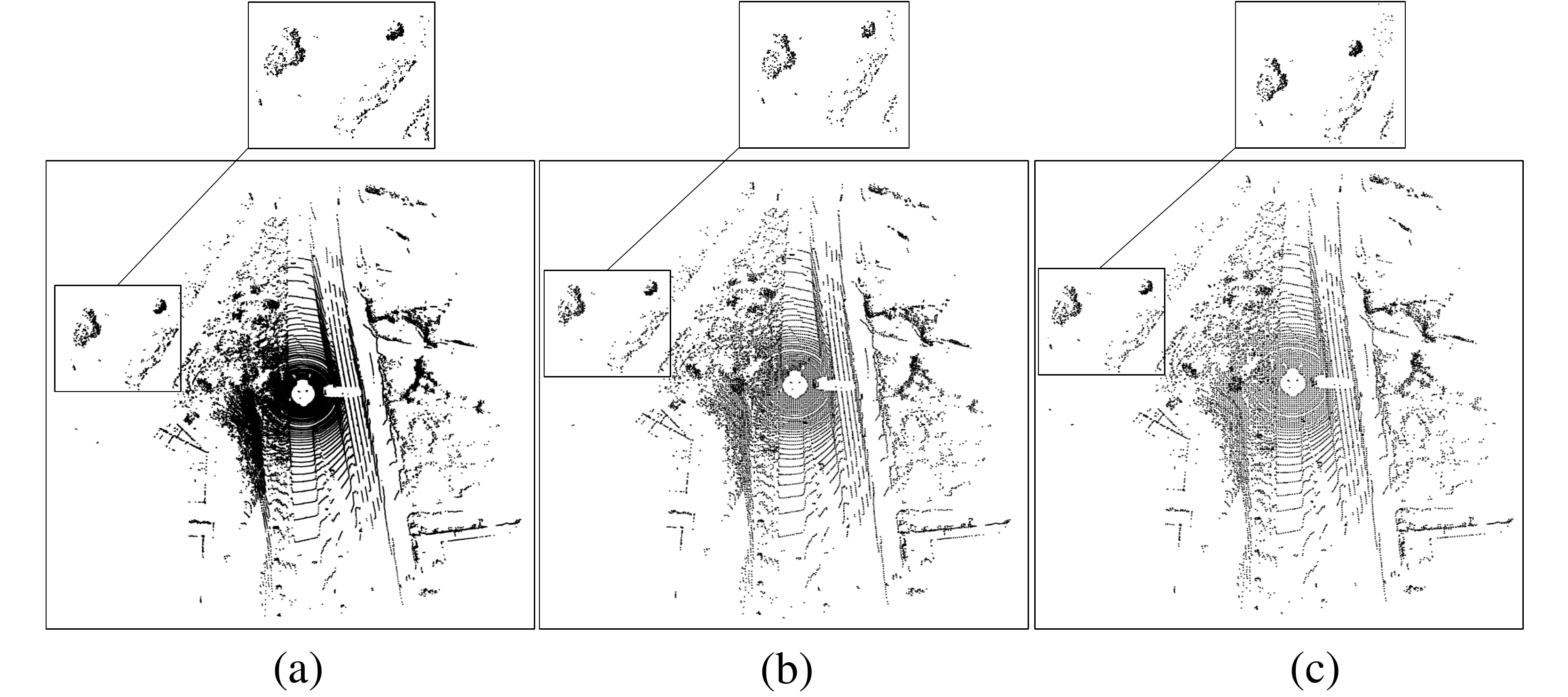}
\end{center}
   \caption{(a). LiDAR scan (b). Uniform downsampling (c). Distance aware downsampling}
   \label{fig:downsample}
\end{figure}

\subsection{Graph Construction}
A point cloud $P$ as consists of $N$ points in $D$ dimensions such that $P = \{p_i| i=1,2,3....n\} \in \mathbb{R}^D$ , where a point $p_i$ is a vector consisting of its coordinates $(x,y,z)$ values, and state values, i.e., reflectance values or encoded features of neighbourhood vertices.

\paragraph{Distance-aware Downsampling} As explained in \cite{shi2020pointgnn}, a single point cloud scan in KITTI dataset commonly comprises tens of thousands of points. It is computationally exorbitant to construct a graph with such a large number of points. Therefore, we introduce a distance-aware downsampling scheme to downsample points $P$ without loosing the relevant information in the original point cloud scan. A simple voxel downsampling uses a regular voxel grid to create a uniformly downsampled point cloud from an input point cloud. The objects located near the centre of scan have dense construction whereas the objects located far from centre are poorly defined. As apparent in Figure \ref{fig:downsample}, an object located far from the ego vehicle is not well defined in a scan. We employ variable voxel sizes depending on the location of the objects from the origin. The points that are located far from the origin uses a smaller voxel size so that the downsampled point cloud do not loose geometrical information from original point cloud scans, since smaller voxel size tends to downsample less number of points than bigger voxel. As evident in Figure \ref{fig:downsample} the quality of distant objects does not deteriorate in distance aware downsapling, even by increasing the voxel size.

The downsampled point cloud $P_D$ can be used to construct a \textit{k}-nearest neighbour graph $G=\{(V,E)\}$ from this downsampled point cloud $P_D$ where $V = \{p_1,p_2,p_3...p_N\}$ and $E$ consists of edges between point $P_i$ to its neighbour vertices within a fixed radius.

\paragraph{Initial Vertex States} Similar to \cite{lang2019pointpillars}, we use a simplified PointNet \cite{pointnet} layer to embed reflective intensity values and relative coordinates. For each point, a linear layer is applied followed by BatchNorm \cite{ioffe2015batch} and ReLU \cite{nair2010rectified} to embed the reflectence intensity and the relative coordinates values. This layer combines point-wise features with a locally aggregated features which allows learning complex features for characterizing local 3D shape information. The resulting features are used as the initial state values of vertices.
This constructed graph is then passed to the GNN for further processing.

\subsection{Attention Based Feature Agrregation}
   Let $S = {s_1,s_2,s_3,...s_N} \in \mathbb{R}^F$ be a set of input features, associated with vertex $u \in V$, at $k-1$ iteration. A single iteration of a Graph Neural Network (GNN) aggregates features from $k$ nodes in a neighborhood $N(u)$ of a given node $u$ such that the updated feature $s{'}$ of vertex $u$ at $(k+1)$ iteration is given by:
   \begin{equation}
   	  s_{u}^{k+1} = \sigma(W_k . (e^k_{uv} , v \in N(u) ),b_k s_u^k)
   \end{equation}
   Here, $W_k$ and $b_k$ are trainable weight and bias matrices and $\sigma$
   is the activation function (e.g. ReLU) to introduce non-linearity. The function $e^k_{uv}$ aggregates features along the edges. This function updates feature and repeats the process in every iteration.

	Inspired from \cite{velivckovic2017graph}, we propose an attention based aggregation method to refine neighbourhood vertex states using weights. The proposed method can handle unordered point cloud sets and size-fluctuating neighbor relationships.
	Let $\alpha$ be the weighting factor (importance) of node $v$’s message to node $u$, which computes the attention coefficients $a_{uv}$ across all the pairs of $u$ in $V$ and $v$ $\in$ $N(u)$. In a standard GNN, $\alpha$ = $ \frac{1}{|N(v)|}$. We define $\alpha_{uv}$  as the byproduct of an attention mechanism $a$, which computes the attention coefficients $e_{uv}$ across pairs of nodes $u$, $v$ based on their messages:
    \begin{equation}
    e_{uv} = a(W_{k}s_{u}^{k-1},W_{k}s_{v}^{k-1}), v \in N(u)
    \end{equation}
    To capture the local structure of objects and dynamically adapt the weights of edges to the similar neighbours, we defined $e_{uv}$ as:
    \begin{equation}
    e_{uv} = a(\delta x_{uv}, \delta s_{uv}), v \in N(u)
    \end{equation}
    where $\delta x_{uv}$ = $x_v$ - $x_u$, the difference between relative coordinates of vertex and $ \delta s_{uv}$ = $ M(s_v)$ - $M(s_u)$, where $M$ is a feature mapping function, i.e, multi-layer neural network. The relative coordinate difference between vertices learns the spatial relationship between $u$ and neighbour $v$. The feature difference between vertex pairs, assigns more weight to the similar neighbors. Both these terms are concatenated and implemented using a multi-layer neural network such that,
    \begin{equation}
    e_{uv} = MLP(\delta x_{uv} || \delta s_{uv})
    \end{equation}
	After handling the different sized vertices from the neighbourhood of $u$, we normalize the $a_{uv}$ coefficients using a softmax function to compare the importance of vertices across different neighbors and calculate $\alpha_{uv}$ such that
	\begin{equation}
	\alpha_{uv} = \frac{\exp(e_{uv}^t)}{\sum_v v \in N(u) \exp(e_{uv}^t)}
	\end{equation}
	where $a_{uv}$ is the attentional weight of vertex $v$ to vertex $u$
	at the $k^{th}$ iteration.
	Therefore, we formulate one iteration of our GNN as:
	\begin{equation}
	s_u^{k} = (\sum_{v \in N(u)} \alpha_{uv}*W_ks_v^{k-1}) + s_{u}^{k-1}	
	\end{equation}
	where * represents the element wise multiplication. We use this final vertex feature to predict both the class and the oriented bounding box of the object.
\begin{figure*}[ht!]
	\begin{center}
		\includegraphics[width=\textwidth]{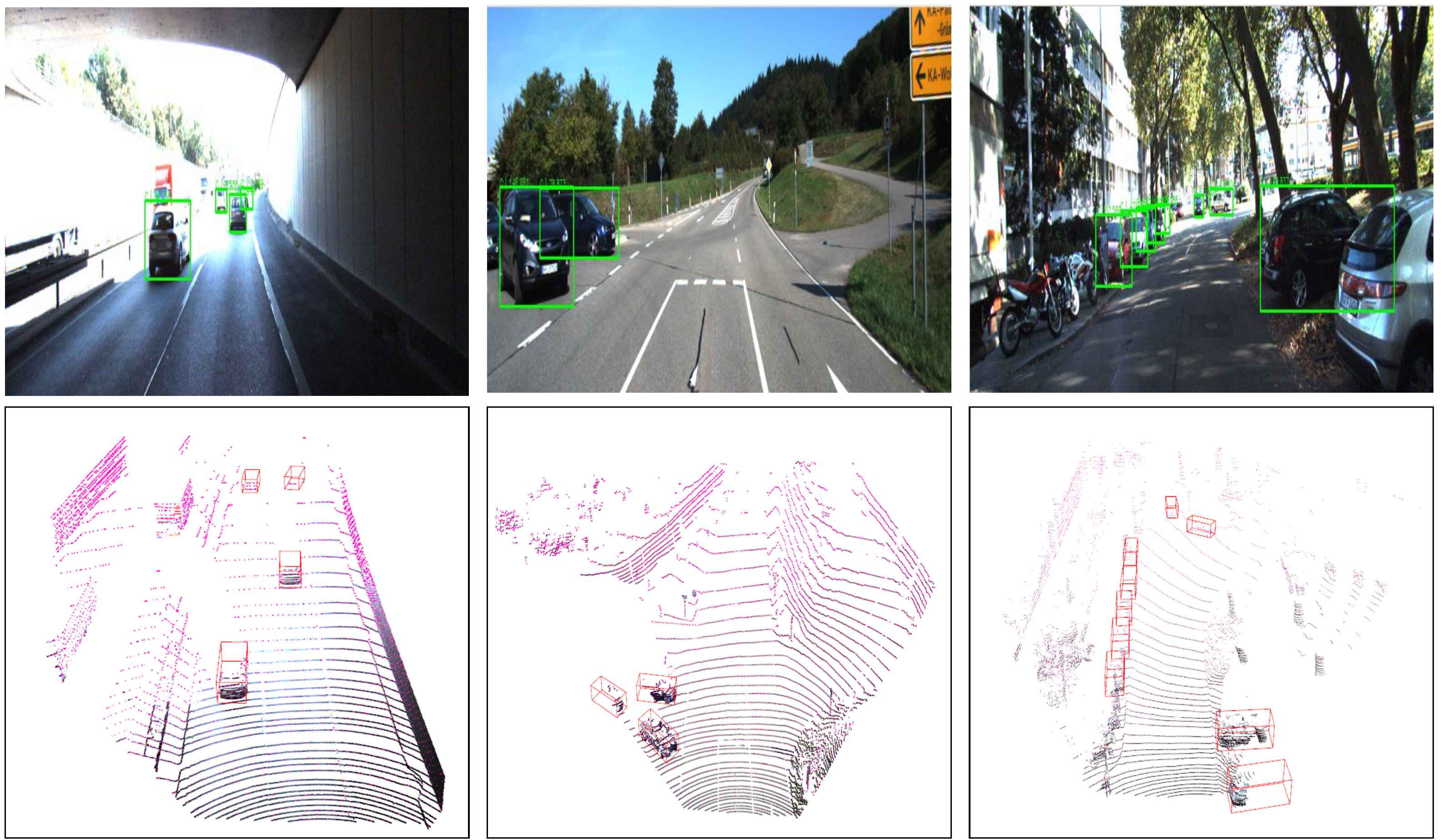}
	\end{center}
	\caption{Qualitative 3D detection results of our architecture on the 'Car' category of KITTI test set. The detected objects
		are shown with red 3D bounding boxes and green 2D bounding boxes. The upper row shows the 2D bounding boxes in the RGB images and the bottom row shows the results in the corresponding point clouds. }
	\label{fig:car}
\end{figure*}
\subsection{Loss}
Our final loss function is composed of three main components, a classification loss, a regression loss and a localization loss. \par
For regression loss $\lambda_{reg}$, we parameterize a 3D ground truth value of a bounding box in seven degrees-of-freedom, such that $b_{gt}$ = ($x_{gt}, y_{gt}, z_{gt},l_{gt},w_{gt},h_{gt}, \theta_{gt}$). Similarly the prior anchor box coordinates are encoded as  $b_a$ = ($x_a, y_a, z_a,l_a,w_a,h_a, \theta_a$). Therefore, the residual difference between the predicted bounding boxes and ground truth boxes is given by:

\begin{multline}
\delta x =  \frac{x_{gt} - x_a}{\delta d}, \delta y =  \frac{y_{gt} - y_a}{\delta d}, \delta z =  \frac{z_{gt} - z_a}{\delta d}
\\
\delta l = \log( \frac{l_{gt}}{l_a}), \delta w = \log( \frac{w_{gt}}{w_a}), \delta h = \log( \frac{h_{gt}}{h_a})
\\
\delta\theta = \sin(\theta_{gt} - \theta_a)
\end{multline}

where $d_a$ is $\sqrt{(w_a)^2 + (l_a)^2}$.
For classification loss $\lambda_{cls}$, we use average cross entropy loss which is given by:
\begin{equation}
	\lambda_{cls} = - \frac{1}{\textit{N}} \sum_{i=1}^{\textit{N}} \sum_{j=1}^{\textit{M}}y_l^i \log(p_l^i)
\end{equation}
where $y_l^i$ is the class label and  $p_l^i$ is the predicted probability.

Similar to \cite{shi2020pointgnn}, we use Huber loss to localize objects belonging to a class we are predicting. All the irrelevant classes are localized as background classes. After that, we average the localization loss of all relevant class objects. The localization loss $\lambda_{loc}$ is given by:
\begin{equation}
\lambda_{loc} =  - \frac{1}{\textit{N}} \sum_{i=1}^{\textit{N}}(v \in b_a) \sum_{\lambda\in\lambda_{gt}} \lambda_{huber}(\lambda b_a - \lambda b_{gt})	
\end{equation}
Therefore the total loss $\lambda_{total}$ is given by:
\begin{equation}
\lambda_{total} = \alpha\lambda_{reg} + \beta\lambda_{cls} + \gamma\lambda_{loc}
\end{equation}
The weighting parameters $\alpha$, $\beta$ and $\gamma$ are used to adjust the relative weights of each loss.

\begin{figure}[h]
	\begin{center}
		\includegraphics[width=\textwidth]{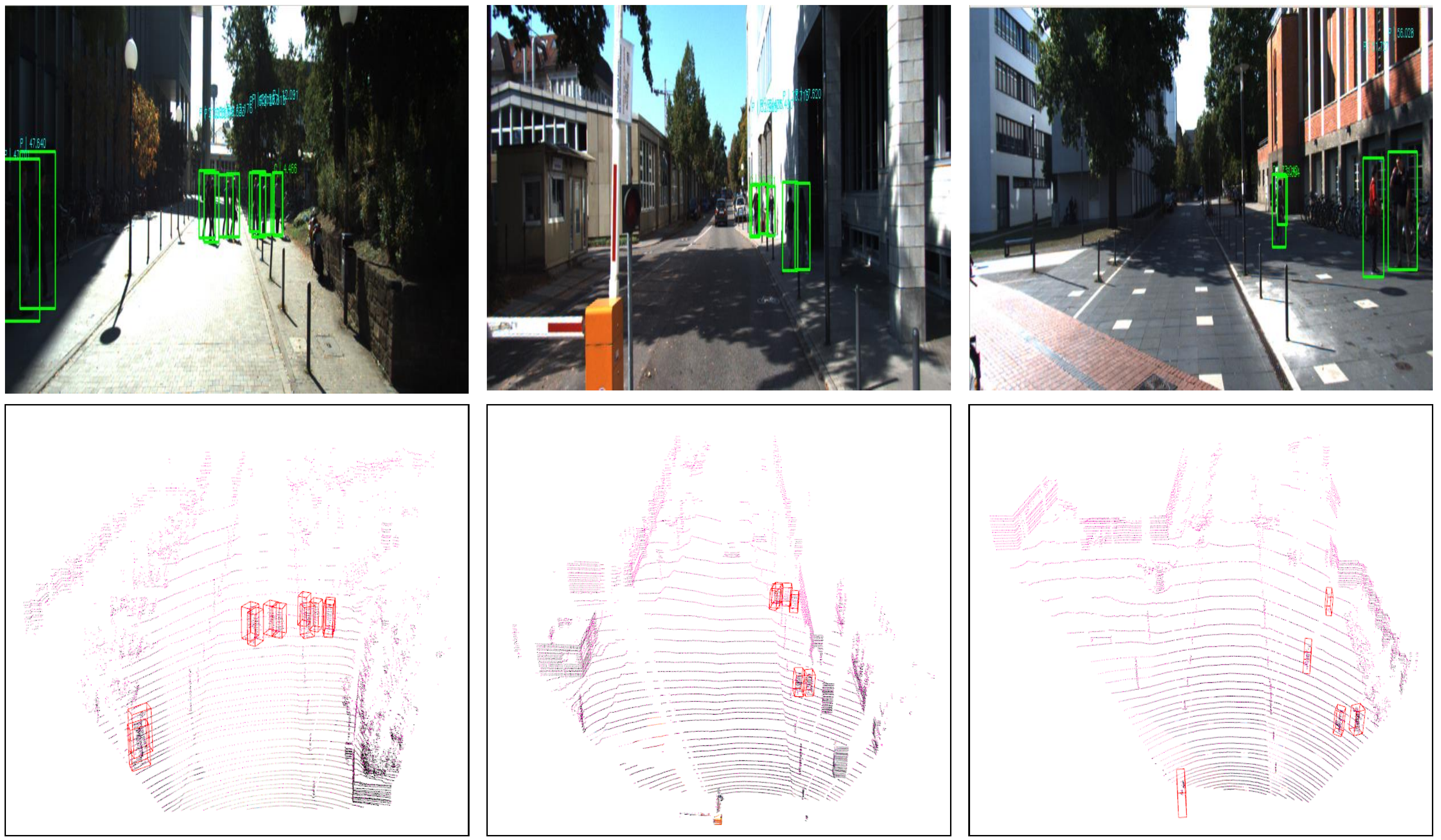}
	\end{center}
	\caption{Qualitative 3D detection results of our architecture on the \enquote*{Pedestrian} category of KITTI test set. The detected objects
		are shown with red 3D bounding boxes and green 2D bounding boxes. The upper row shows the 2D bounding box in the RGB images and the bottom row shows the results in the corresponding point clouds. }
	\label{fig:pedestrian}
\end{figure}
\begin{figure}[h]
	\begin{center}
		\includegraphics[width=\textwidth]{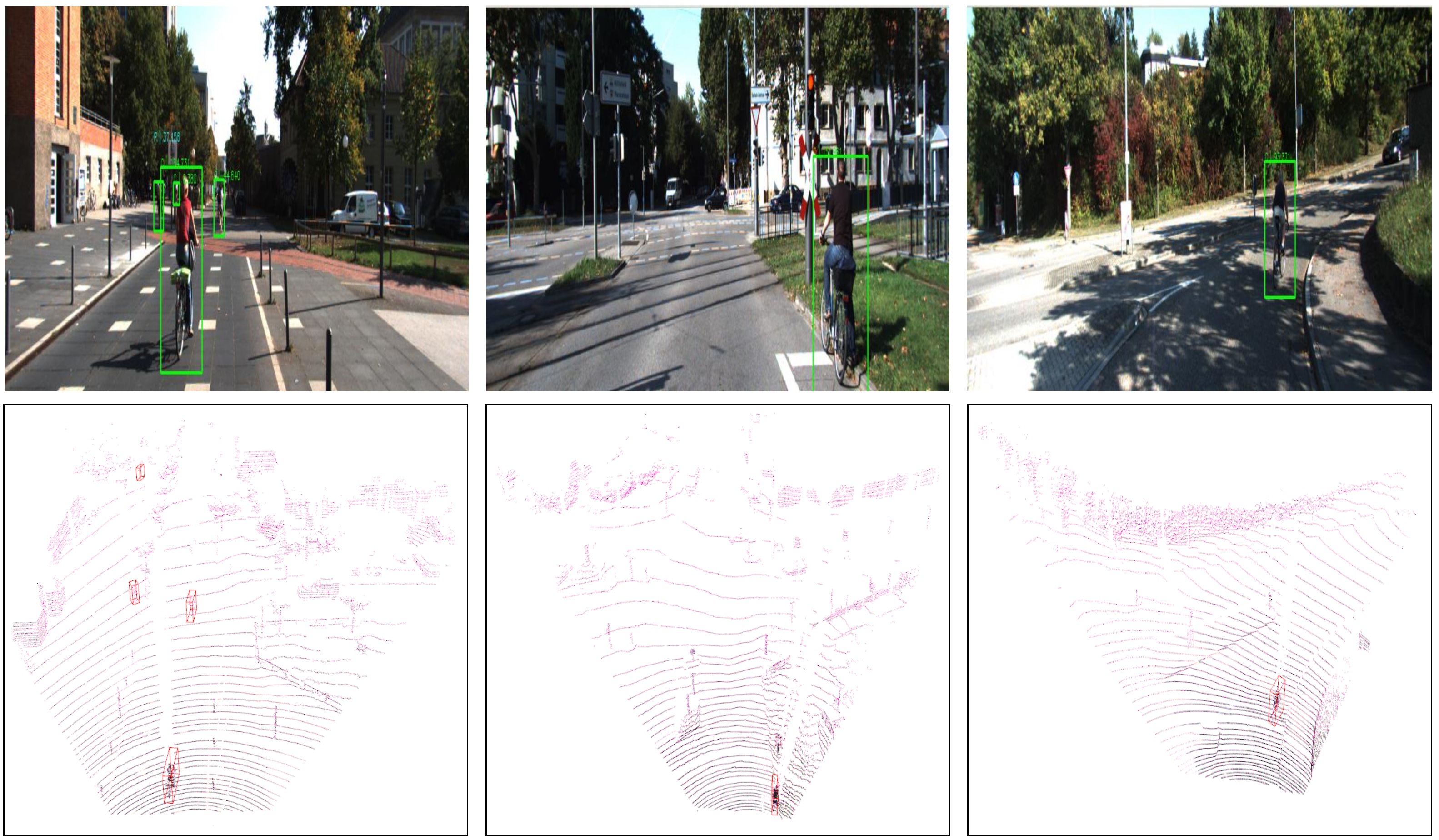}
	\end{center}
	\caption{Qualitative 3D detection results of our architecture on the \enquote*{Cyclist} category of KITTI test set. The detected objects
		are shown with red 3D bounding boxes and green 2D bounding boxes. The upper row shows the 2D bounding box in the RGB images and the bottom row shows the results in the corresponding point clouds. }
	\label{fig:cycle}
\end{figure}

\section{Experiments}
\subsection{Dataset}
We evaluate our architecture on the widely used KITTI object detection benchmark\cite{Geiger2012CVPR} which contains 7481 training samples and 7518 testing samples. Each sample has a point cloud scan, a respective image and calibration data. Since the dataset only annotates objects that are visible within the image, we process the point cloud only within the field of view of the image. Due to the scale differences, we trained the network separately on car, pedestrian and cyclist data.

\begin{table*}[h]
	\small
	\begin{tabular}{ccccccccccc}
		\hline
		Method         & Modality      & \multicolumn{3}{c}{Car}  & \multicolumn{3}{c}{Pedestrian} & \multicolumn{3}{c}{Cyclist}                  \\
		\multicolumn{2}{c}{}           & Easy  & Moderate & Hard  & Easy    & Moderate   & Hard    & Easy  & Moderate & Hard \\ \hline
		MV3D \cite{chen2017multi}          & LiDAR + Image & 68.35 & 54.54    & 49.16 & N/A     & N/A        & N/A     & N/A   & N/A      & N/A                       \\
		AVOD \cite{ku2018joint}          & LiDAR + Image & 76.39 & 66.47    & 60.23 & 36.10   & 27.86      & 25.76   & 57.19 & 42.08    & 38.29                     \\
		F-PointNet \cite{qi2018frustum}     & LiDAR + Image & 81.20 & 70.39    & 62.19 & 51.21   & 44.89      & 40.23   & 72.27 & 56.12    & 49.01                     \\
		UberATG-MMF \cite{liang2019CVPR}   & LiDAR + Image & 86.81 & 76.75    & 68.41 & N/A     & N/A        & N/A     & N/A   & N/A      & N/A                       \\ \hline
		VoxelNet \cite{zhou2018voxelnet} & LiDAR         & 77.47 & 65.11    & 57.73 & 39.48     & 33.69        & 31.51     & 61.22   & 48.36      & 44.37  \\
		Pseudo-LiDAR++ \cite{you2020pseudolidar} & LiDAR         & 61.11 & 42.43    & 36.99 & N/A     & N/A        & N/A     & N/A   & N/A      & N/A                       \\
		BirdNet+ \cite{Barrera2020}       & LiDAR         & 70.14 & 51.85    & 50.03 & 37.99   & 31.46      & 29.46   & 67.38 & 47.72    & 42.89                     \\
		DSGN \cite{Chen2020dsgn}           & LiDAR         & 73.50 & 52.18    & 45.14 & 20.53   & 15.55      & 14.55   & 27.76 & 18.17    & 16.21                     \\
		CG-Stereo \cite{li2020confidence}     & LiDAR         & 74.39 & 53.58    & 46.50 & 33.22   & 24.31      & 20.95   & 47.40 & 30.89    & 27.23                     \\
		PointGNN \cite{shi2020pointgnn}       & LiDAR         & 88.33 & 79.47    & 72.29 & 51.92   & 43.77      & 40.14   & 78.60 & 63.40    & 57.08                     \\ \hline
		Ours           & LiDAR         & 75.67 & 63.90    & 55.09 & 43.62   & 34.56      & 31.34   & 58.44 & 41.81    & 36.69 \\ \hline
	\end{tabular}
	\caption{\label{tab:3DAP}The Average Precision (AP) comparison of 3D object detection on the KITTI test dataset.}
\end{table*}

\begin{table*}[h]
	\small
	\begin{tabular}{ccccccccccc}
		\hline
		Method         & Modality      & \multicolumn{3}{c}{Car}  & \multicolumn{3}{c}{Pedestrian} & \multicolumn{3}{c}{Cyclist}                  \\
		\multicolumn{2}{c}{}           & Easy  & Moderate & Hard  & Easy    & Moderate   & Hard    & Easy  & Moderate & Hard \\ \hline
		MV3D \cite{chen2017multi}          & LiDAR + Image & 86.62 & 78.93    & 69.80 & N/A     & N/A        & N/A     & N/A   & N/A      & N/A                       \\
		AVOD \cite{ku2018joint}          & LiDAR + Image & 89.75 & 84.95    & 78.32 & 42.58   & 33.57      & 30.14   & 64.11 & 48.15    & 42.37                     \\
		F-PointNet \cite{qi2018frustum}     & LiDAR + Image & 91.97 & 84.67    & 74.77 & 57.13   & 49.57      & 45.48   & 77.26 & 61.37    & 53.78                     \\
		UberATG-MMF \cite{liang2019CVPR}   & LiDAR + Image & 93.67 & 88.21    & 81.99 & N/A     & N/A        & N/A     & N/A   & N/A      & N/A                       \\ \hline
		VoxelNet \cite{zhou2018voxelnet} & LiDAR         & 89.35 & 79.26    & 77.39 & 46.13     & 40.74        & 38.11     & 66.70   & 54.76      & 50.55  \\
		Pseudo-LiDAR++ \cite{you2020pseudolidar} & LiDAR         & 78.31 & 58.01    & 51.25 & N/A     & N/A        & N/A     & N/A   & N/A      & N/A                       \\
		BirdNet+ \cite{Barrera2020}       & LiDAR         & 84.80 & 63.33    & 61.23 & 45.53   & 38.28      & 35.37   & 72.45 & 52.15    & 46.57                     \\
		DSGN \cite{Chen2020dsgn}           & LiDAR         & 82.90 & 65.05    & 56.60 & 26.61   & 20.75      & 18.86   & 31.23 & 21.04    & 18.23                     \\
		CG-Stereo \cite{li2020confidence}     & LiDAR         & 85.29 & 66.44    & 58.95 & 39.42   & 29.56      & 25.87   & 55.33 & 36.25    & 32.17                     \\
		PointGNN \cite{shi2020pointgnn}       & LiDAR         & 93.11 & 89.17    & 83.90 & 55.36   & 47.07      & 44.61   & 81.17 & 67.28    & 59.67                     \\ \hline
		Ours           & LiDAR         & 87.95 & 80.65    & 70.97 & 48.26   & 39.41      & 35.90   & 66.56 & 47.80    & 41.82    \\ \hline
	\end{tabular}
	
	\caption{\label{tab:BEVAP} The Average Precision (AP) comparison of Bird’s Eye View (BEV) object detection on the KITTI test dataset.}
\end{table*}
\begin{figure*}[ht!]
	\begin{center}
		\includegraphics[width=\textwidth]{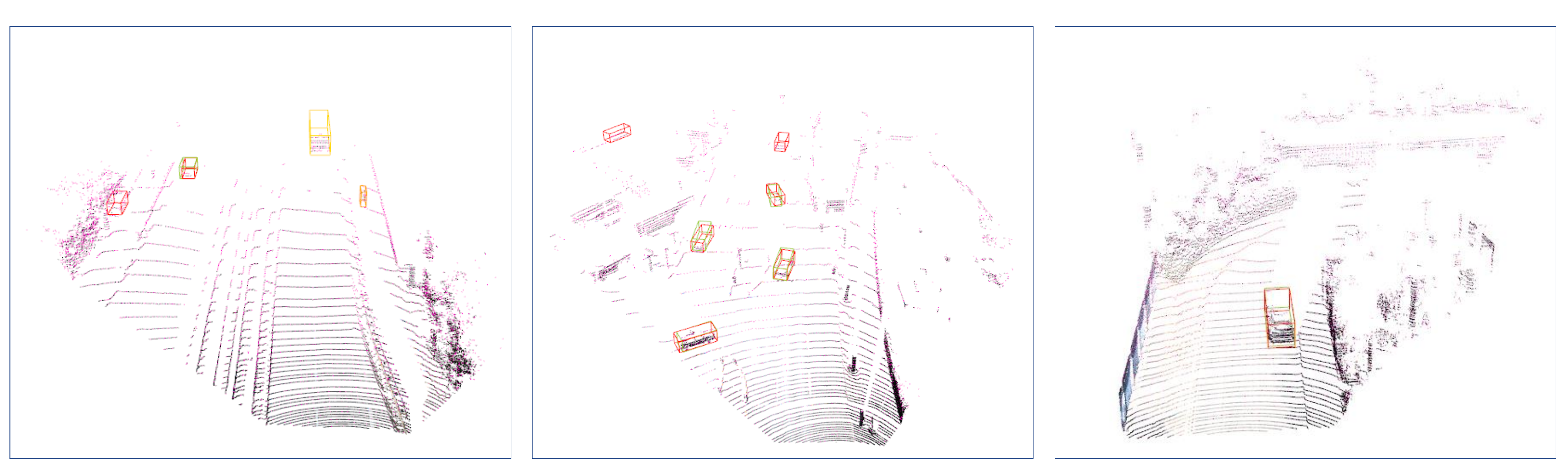}
	\end{center}
	\caption{Qualitative 3D detection results of our architecture on the KITTI validation set. The detected objects
		are shown with red 3D bounding boxes and green 3D bounding boxes reflect the ground truth bounding boxes. }
	\label{fig:car3D}
\end{figure*}

\subsection{Implementation Details}
We train the proposed network end-to-end with a batch size of 2. The loss weights are $\alpha$= 0.1, $\beta$= 10, $\gamma$= 0.0005.. For car, we use an initial learning rate of 0.125 and a decay rate of 0.1 every 400K steps. We trained the network for 1400K steps. For pedestrian, we used a learning rate of 0.25 and a decay rate of 0.25 every 400K steps and trained for 1000K steps. Similarly for cyclist, we trained for 1000K steps with a learning rate of 0.32 and a decay rate of 0.25 after every 400k steps.

For cars, we use the anchor box size of (1.6, 3.9, 1.5) meters covering width, length and height respectively with two rotations 0 and 90 degrees. Similar to \cite{shi2020pointgnn} we set \textit{r} to 1.8 m. The anchor box sizes for pedestrian and cyclist objects were set to (0.6, 0.8, 1.73) meters and (0.6, 1.76, 1.73) meters respectively. We used variable voxel sizes during the sampling, i.e.,  we used voxel sizes of 0.5 and 0.8 for points greater than 40 m and less than 20 m respectively from the sensor along the z axis. To reduce redundancy, we apply IoU threshold of 0.7 for NMS for car category and 0.6 threshold for pedestrian and cyclist category. The network was trained in an end-to-end manner on a single TITAN V GPU. We employed the ADAM optimizer to train our network. Furthermore, we also applied data augmentation as in \cite{lang2019pointpillars} to prevent overfitting and make predictions robust.

\subsection{Qualitative Results}
We illustrate our prediction results on KITTI test data in Figures \ref{fig:car}-\ref{fig:cycle}. We visualize 3D bounding boxes in LiDAR scan and 2D bounding boxes on RGB images. From the figures, we can observe that the proposed architecture can estimate accurate 3D bounding boxes in a variety of scenes. The architecture can predict the correct positions in poor lighting conditions and occlusions. As shown in Figure \ref{fig:pedestrian}, one can see that the model is capable of predicting the pedestrian positions even when they are not clearly visible in RGB images.

\subsection{Quantitative Comparison}
We validate the proposed approach using a set of experiments carried out on the KITTI object detection benchmark. In Table \ref{tab:3DAP} and Table \ref{tab:BEVAP}, we report the accuracy results of our method in terms of the average precision (AP) values on 3D and bird's eye view (BEV) datasets respectively, along with the accuracy results of various other 3D detection methods. The KITTI dataset evaluates the average precision (AP) on three difficulty levels: easy, moderate, and hard. The 3D and BEV detection results obtained by our proposed approach are comparable to the ones provided by the other state-of-the-art methods. Our approach detects all three classes in KITTI dataset reasonably well despite not achieving the top score. The proposed model even outperform fusion based MV3D in car category, and AVOD in pedestrian category (refer to Table \ref{tab:3DAP}).

\subsection{Ablation Study}
In this section, we discuss a series of experiments conducted on  validation subset of KITTI dataset to illustrate the role of each module and the different hyperparameters in improving the final results. Similar to \cite{lang2019pointpillars, yan2018second, shi2020pointgnn}, we split the KITTI dataset into 3712 training samples and 3769 validation samples. We follow the approach used in KITTI benchmark, and calculate 3D AP and BEV accuracy.  All our ablation studies are experimented on the car class since it  contains the largest number of training samples.

\textbf{Results on KITTI Validation Dataset:} We test our method on the car category of KITTI validation dataset and compare our results with state-of-art methods. The result comparison on 3D AP and BEV is shown in Table \ref{tab:3DAPVAL} and Table \ref{tab:BEVAPVAL} respectively. For car category, the proposed method achieves results comparable to state-of-the-art methods on all the difficulty levels of the KITTI dataset. We visualize our results on KITTI validation split (refer to Figure \ref{fig:car3D}).
 \begin{table}[h]
 \begin{tabular}{cccc}
 	\hline
 	Method     &       & Car      &       \\ \hline
 	& Easy  & Moderate & Hard  \\ \hline
 	MV3D       & 71.29 & 62.68    & 56.56 \\
 	AVOD       & 84.41 & 74.44    & 68.65 \\
 	F-Pointnet & 83.76 & 70.92    & 63.65 \\
 	DSGN       & 72.31 & 54.27    & 47.71 \\
 	Voxelnet   & 81.97 & 65.46    & 62.85 \\
 	PointGNN   & 87.89 & 78.34    & 77.38 \\ \hline
 	Ours       & 83.54 & 74.47    & 63.84 \\ \hline
 \end{tabular}
 \caption{\label{tab:3DAPVAL} The Average Precision (AP) comparison of 3D object detection on the KITTI validation dataset.}
 \end{table}
 \begin{table}[h]
	\begin{tabular}{cccc}
		\hline
		Method     &       & Car      &       \\ \hline
		& Easy  & Moderate & Hard  \\ \hline
		MV3D       & 86.55 & 78.10    & 76.67 \\
		F-Pointnet & 88.16 & 84.02    & 76.44 \\
		DSGN       & 83.24 & 63.91    & 57.83 \\
		Voxelnet   & 89.60 & 84.81    & 78.57 \\
		PointGNN   & 89.82 & 88.31    & 87.16 \\ \hline
		Ours       & 90.12 & 87.05    & 75.48 \\ \hline
	\end{tabular}
	\caption{\label{tab:BEVAPVAL} The Average Precision (AP) comparison of BEV detection on the KITTI validation dataset.}
\end{table}

\textbf{Effects of Different Number of Layers:} In our architecture, we stack $n$ number of GNN layers to extract aggregated features. To demonstrate the influence of changing the value of $n$, we train our network with $n$ varying from
1 to 4. We demonstrate our results in Table \ref{tab:3DPERF}. Table \ref{tab:3DPERF} indicates there's a slight increase in the accuracy when $n$ is changed from 1 to 3, which can be attributed to the fact that the neighbourhood features are being aggregated to the vertex itself. Our model performance continues to increase as we increase the value of $n$. There is a slight decrease in accuracy at $n=4$, which indicates that our neural network might be overtrained.
\begin{table}[h]
	\begin{tabular}{cccc}
		\hline
		Number of layers (n) &       & Car (3D AP\%) &       \\ \hline
		& Easy  & Moderate      & Hard  \\ \hline
		1                    & 80.24 & 72.27         & 62.78 \\
		2                    & 82.73 & 73.65         & 63.14 \\
		3                    & 83.54 & 74.47         & 63.84 \\
		4                    & 83.14 & 74.21         & 63.46 \\ \hline
	\end{tabular}
\caption{\label{tab:3DPERF} The 3D Average Precision (AP) comparison when changing the number of layers of our proposed GNN.}
\end{table}

\textbf{Inference Time Analysis}: For any algorithm to be deployed in autonomus vehicle scenario, the inference speed plays a crucial role. The algorithm must be able to predict the oncoming object's position in real-time. The performance of an algorithm is subject to the hardware and code-optimization. Our architecture is written in Python, and implemented in Tensorflow for GPU computation. We measure our inference time on a machine with Intel i7-8700k CPU, 32GB RAM and Nvidia Titan V GPU. The dataset reading and preprocessing takes 45 ms. The nearest neighbour graph construction consumes 112 ms. A single iteration of our GNN takes 270 ms and it takes 15 ms for final bounding box predictions.
\section{Conclusion}
In this work, we proposed dynamic edge weight based graph neural network (GNN) for 3D object detection for autonomous driving applications. The proposed architecture is different from the state-of-the-art by using a graph representation of point cloud data rather than the volumetric based representation.  Furthermore, the proposed architecture employs attention technique for feature extraction formulated on the relative coordinates and features similarity between two vertices. Our architecture is found to produce accurate bounding box positions in road scenes. Experiments on the KITTI dataset benchmark show the acceptable results of our proposed architecture 3D classification, bounding box location estimation, and category classification on bird's eye view benchmark. Finally, the proposed architecture produces results in low inference times. Future work will be to improve the current prediction results and code-optimization to improve better inference speeds.


{\small
\bibliographystyle{ieee}
\bibliography{egbib}
}

\end{document}